\pdfoutput=1
\documentclass{article}

    \PassOptionsToPackage{numbers, compress}{natbib}

\usepackage[final]{neurips_2024}

\usepackage[utf8]{inputenc} 
\usepackage[T1]{fontenc}    
\usepackage{hyperref}       
\usepackage{url}            
\usepackage{booktabs}       
\usepackage{amsfonts}       
\usepackage{nicefrac}       
\usepackage{microtype}      
\usepackage{xcolor}         
\usepackage{ulem}           
\usepackage{graphicx}
\usepackage{wrapfig}
\usepackage{algorithm}
\usepackage{algorithmic}
\usepackage{amsmath}
\usepackage{multirow}
\usepackage{array} 
\usepackage[numbers]{natbib}

\title{V“Mean”ba: Visual State Space Models only need 1 hidden dimension}

\author{%
  Tien-Yu Chi \\
  National Yang Ming Chiao Tung University \\
  \texttt{b03902059@ntu.edu.tw}
  \And
  Hung-Yueh Chiang \\
  The University of Texas at Austin \\
  \texttt{hungyueh.chiang@utexas.edu}
  \And
  Chi-Chih Chang \\
  Cornell University \\
  \texttt{cc2869@cornell.edu}
  \And
  Ning-Chi Huang \\
  National Yang Ming Chiao Tung University \\
  \texttt{nchuang@cs.nycu.edu.tw}
  \And
  Kai-Chiang Wu \\
  National Yang Ming Chiao Tung University \\
  \texttt{kcw@cs.nctu.edu.tw}
}

\begin{document}

\maketitle

\begin{abstract}
Vision transformers dominate image processing tasks due to their superior performance. However, the quadratic complexity of self-attention limits the scalability of these systems and their deployment on resource-constrained devices. State Space Models (SSMs) have emerged as a solution by introducing a linear recurrence mechanism, which reduces the complexity of sequence modeling from quadratic to linear. Recently, SSMs have been extended to high-resolution vision tasks. Nonetheless, the linear recurrence mechanism struggles to fully utilize matrix multiplication units on modern hardware, resulting in a computational bottleneck. We address this issue by introducing \textit{VMeanba}, a training-free compression method that eliminates the channel dimension in SSMs using mean operations. Our key observation is that the output activations of SSM blocks exhibit low variances across channels. Our \textit{VMeanba} leverages this property to optimize computation by averaging activation maps across the channel to reduce the computational overhead without compromising accuracy.
Evaluations on image classification and semantic segmentation tasks demonstrate that \textit{VMeanba} achieves up to a 1.12x speedup with less than a 3\% accuracy loss. When combined with 40\% unstructured pruning, the accuracy drop remains under 3\%.
\end{abstract}

\section{Introduction}\label{section:introduction}

Computer vision has advanced significantly due to deep learning and the availability of large-scale datasets. Convolutional Neural Networks (CNNs) have become foundational for tasks such as image classification \cite{AlexNet2012,vgg,he2016deep} and object detection \cite{girshick2014rich,girshick2015fast,redmon2016you}. However, CNNs struggle to capture long-range dependencies. Vision Transformers (ViTs) \cite{ViT2021,Swin2021,DeiT2021} which utilize self-attention mechanisms, effectively address this limitation but suffer from high computational costs due to quadratic complexity. To mitigate these costs, research has focused on reducing ViT complexity \cite{wang2020linformer, beltagy2020longformer, Swin2021, liu2022swin, liu2023efficientvit}, applying model compression techniques \cite{liu2021post,lin2021fq, zhu2021vision,yang2021nvit, DeiT2021, lin2023supervised}, and exploring alternative architectures like RWKV and State Space Models (SSMs) \cite{rmkvpeng2023rwkv, s4gu2021, h3fu2022hungry, gu_mamba_2024}.

State Space Models (SSMs) have recently garnered attention in computer vision as efficient and effective alternatives to Vision Transformers (ViTs), demonstrating competitive performance across various tasks \cite{liu_vmamba_2024,zhu2024vision,li2024videomamba,teng2024dim}. For example, VMamba \cite{liu_vmamba_2024} achievesd 82.6\% top-1 accuracy on ImageNet-1k \cite{deng2009imagenet}, surpassing Swin Transformer \cite{Swin2021} by 1.3\% with comparable FLOPs. However, despite reducing computational complexity,  SSMs still fail to fully utilize matrix multiplication units on GPUs, creating a bottleneck in vision-based SSM models.

\begin{wrapfigure}{i}{0.5\textwidth} 
    \centering
    \includegraphics[width=0.45\textwidth]{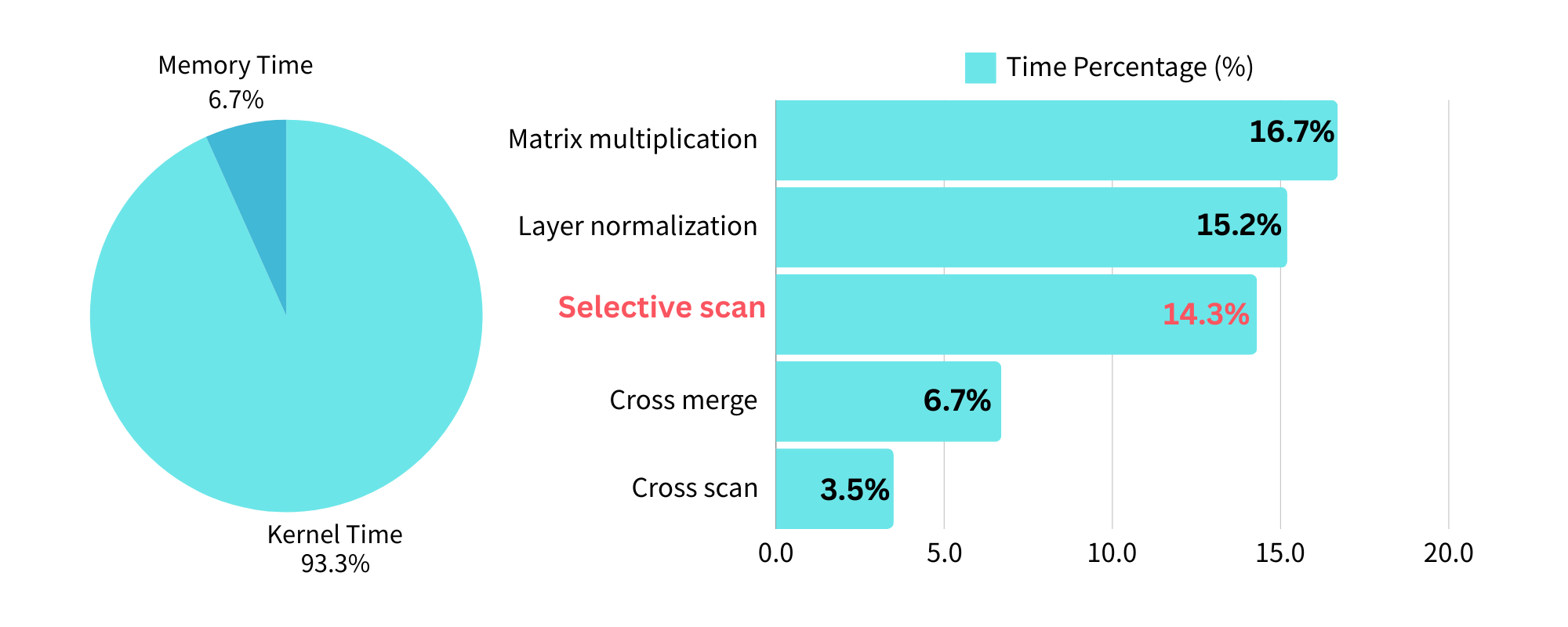}
    \caption{
    The GPU kernel time of each operation in a VMamba block. The latency is measured using feature maps with an input resolution of $224 \times 224$. We rank the kernels by their latency and highlights the top-5 time-consuming kernels on the bar chart. The selective scan operation is one of the major contributors in the VMamba block.
    }
    \label{fig1}
\end{wrapfigure}

To this end, we first analyze the latency breakdown of VMamba \cite{liu_vmamba_2024} and identify the selective scan operation \cite{gu_mamba_2024} as one of the key bottlenecks in inference. Figure \ref{fig1} shows that the selective scan accounts for 14.3\% of the total kernel time in a VMamba block. While optimizing selective scan operation is critical for enhancing SSM efficiency, few research works address this problem and optimizing the efficiency of SSMs remains unexplored.

\begin{figure*}[h!]
    \centering
    \includegraphics[width=0.8\textwidth]{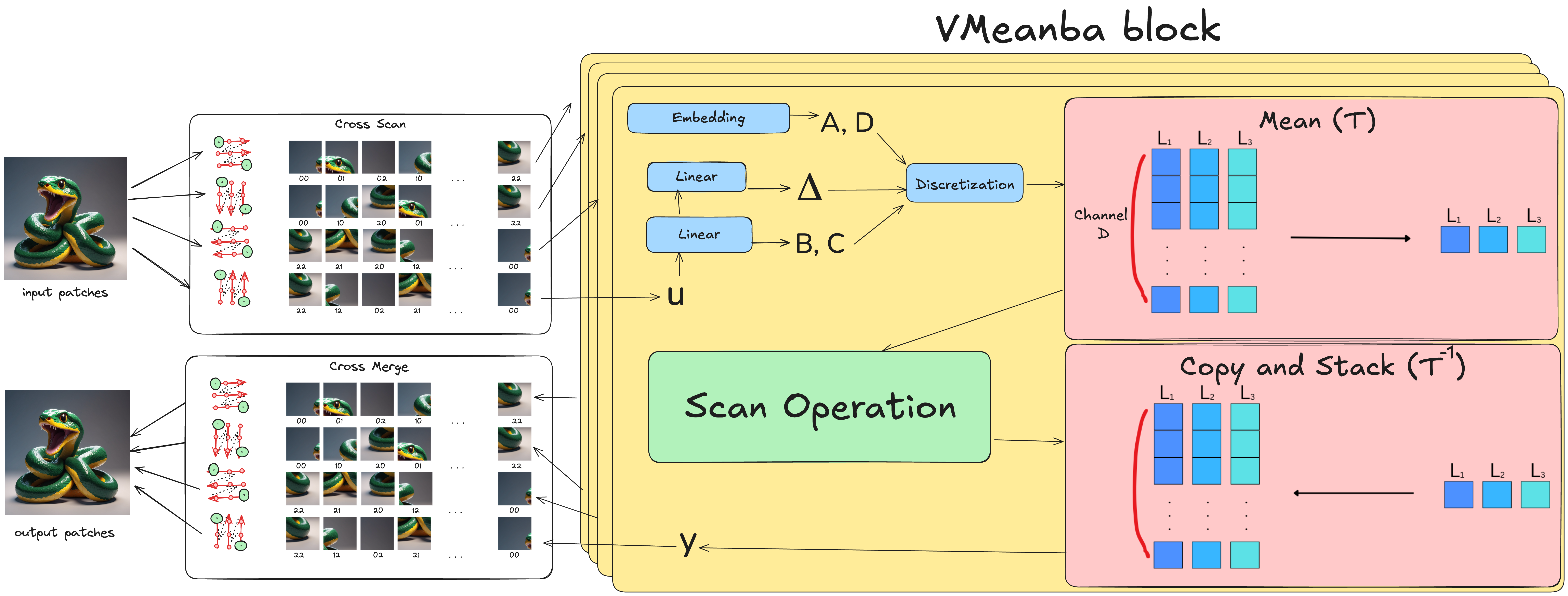} 
    \caption{Overview of the \textit{VMeanba} block. \textit{VMeanba} reduced the channel dimension of the inputs to the associated scan operation by applying a transform $T$, thereby simplifying the computation. The proposed \textit{VMeanba} components are highlighted in red, while the original selective scan components are shown in blue and green, with the green block indicating the main area of optimization.} 
    \label{fig2}
\end{figure*}

In this paper, we propose \textit{VMeanba}, a novel activation compression method designed to optimize the selective scan operation in VMamba blocks. The high-level overview of \textit{VMeanba} is presented in Figure \ref{fig2}. The key idea is to reduce the input tensor's channel dimensions in the associate scan operation by applying a mean operation. Through analysis of the weight and activation distributions in the trained VMamba model, we identified a smooth pattern with small variances that allows for dimensional reduction. Based on this observation, we developed the \textit{VMeanba} block to exploit this pattern, resulting in a more efficient associate scan operation without compromising accuracy. Experimental results demonstrate that \textit{VMeanba} achieves up to a 1.12x speedup with less than a 3\% accuracy loss. To the best of our knowledge, this is the first work optimizing of the selective scan operation in VMamba.

\section{Methods} \label{section:method}
\subsection{Distribution Analysis of VMamba}\label{subsection:vmamba}
\begin{figure}
    \centering
    \includegraphics[width=1\textwidth]{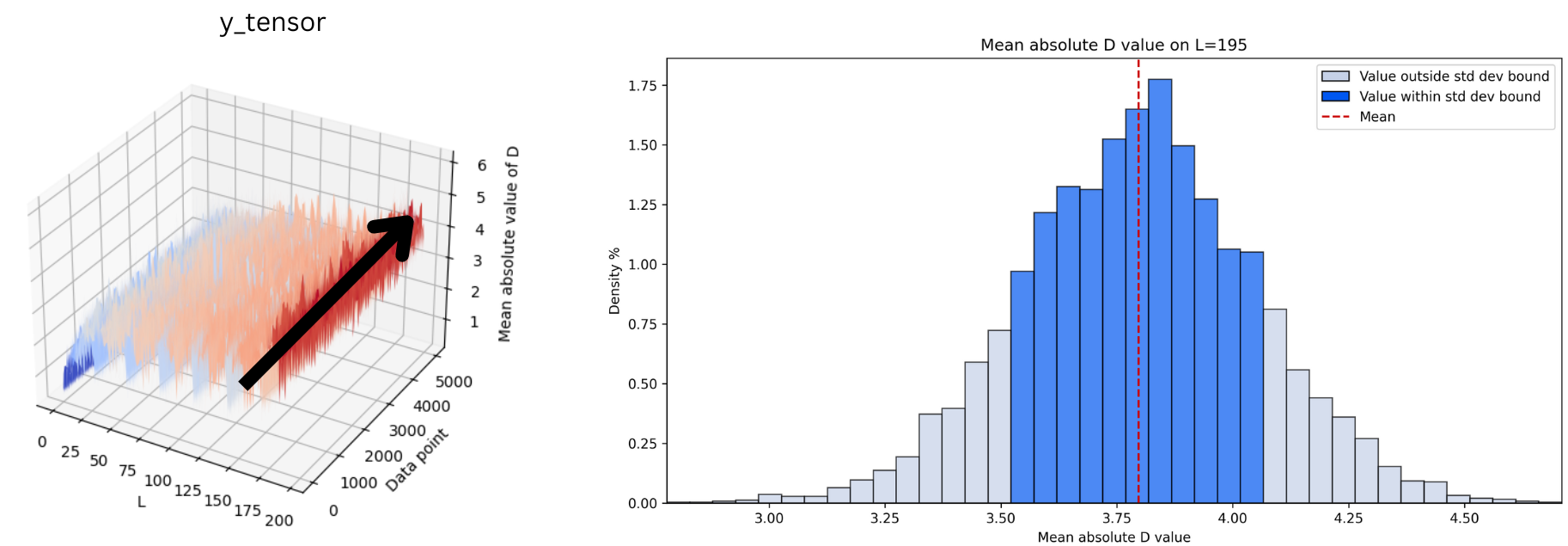} 
    \caption{The figure illustrates the distribution of inner dimension values of $y_{layer}$ across various data points as a function of sequence length. Notably, the distribution remains remarkably consistent across different data points for identical $l$ values, as indicated by the arrows. The distribution for $l=195$ , shown on the right, provides further evidence of this concentration.
    }
    \label{fig3}
\end{figure}

We conduct an in-depth investigation into the characteristics of each layer's output within the Mamba block of VMamba. The output is denoted as $y_{layer} \in \mathbb{R}^{B \times D \times L}$, where $B$ is the batch size, $D$ is the inner channel dimension utilized by the scan algorithm within the Mamba block, and $L$ is 4x of the feature map size $HW$. Our analysis revealed that for each $y_{layer}$, the distribution of values across the inner channel dimension is remarkably consistent across different data points, as illustrated in figure \ref{fig3}. This observation raised a critical question:  Is the full dimensionality $D$ necessary for each $y_{layer}$?

To explore this, we hypothesized a property described by equation (\ref{eq:7}):
\begin{equation}\label{eq:7}
    \begin{gathered}
        y_{\text{layer}}[:, d, :] \sim y_{layer}[:, d', :] \quad \\
        \forall d, d' \in [1, D], d \neq d'
    \end{gathered}
\end{equation}
Given that the scan algorithm in the Mamba block performs a linear transformation, this unique property of $y_{layer}$ can be attributed to the inputs $\bar{A}$, $\bar{B}u_t$ and $C$ to the SSM system. Consequently, we propose that a reduced set of inputs $(\bar{A}_{reduce}, (\bar{B}u_t)_{reduce}, C_{reduce})$, referred to collectively as $I_{basis}$, can effectively represent the original inputs $(\bar{A}, \bar{B}u_t, C)$. By leveraging these reduced inputs, we can optimize the computational efficiency of each Mamba block. 

\subsection{VMeanba}\label{subsection:vmeanba}
Building on the findings from section \ref{subsection:vmamba}, we indroduce a new model inference efficiency optimization method called \textit{VMeanba}, which computes $I_{basis}$ for each Mamba block using mean
 operators. We further design a pipeline to select which layers in the model will undergo this optimization.\\
\textbf{VMeanba block.} The $I_{basis}$ is derived by having a transform function $T$ that maps the original inputs $(\bar{A}, \bar{B}u_t, C)$ to reduced dimension inputs. After processing by the 
original Mamba block, the output is recovered using an inverse transform function $T^{-1}$. This entire process can be expressed as equation (\ref{eq:9}).
\begin{equation}\label{eq:9}
    \begin{gathered}
        y_{layer} = T^{-1}(Mamba(T(\bar{A}, \bar{B}u_t, C)))
    \end{gathered}
\end{equation}
In this process, $T$ is defined as the mean operator applied along the inner channel dimension axis, and $T^{-1}$ is defined as the broadcast operator. While the mean transform may lead to a loss of information, it significantly reduces the dimensionality of the inputs from $D$ to $1$, with our experiments demonstrating that model performance is maintained. The computational complexity analysis is provided in \ref{appendix:computation-complexity}.\\
\textbf{Layer Selection.} We developed a pipeline to replace $K$ Mamba blocks with \textit{VMeanba} blocks. We treat the choices of layers as a hyperparameter, determined using the validation set. Specifically, we calculate the layer impact score $S_{layer}$ for each layer, and select the layers with the $K$ 
 smallest scores to apply the VMeanba optimization.
The impact score is defined by equation (\ref{eq:10}):
\begin{equation}\label{eq:10}
    \begin{aligned}
        S_{layer} = Acc(Original Model) - Acc(VMeanba\ on \ layer)
    \end{aligned}
\end{equation}
where $Acc$ represents the model accuracy on the validation set. The algorithm for this process is detailed in \ref{appendix:algorithm}.
\section{Experiments}\label{section:experiments}
We apply the proposed \textit{VMeanba} method to two different tasks: image classification and semantic segmentation. 
The details experiment setup and more experiments are provided in appendix \ref{appendix:exp-setup}, \ref{appendix:more-exp}

\begin{figure*}
    \centering
    \includegraphics[width=1\textwidth]{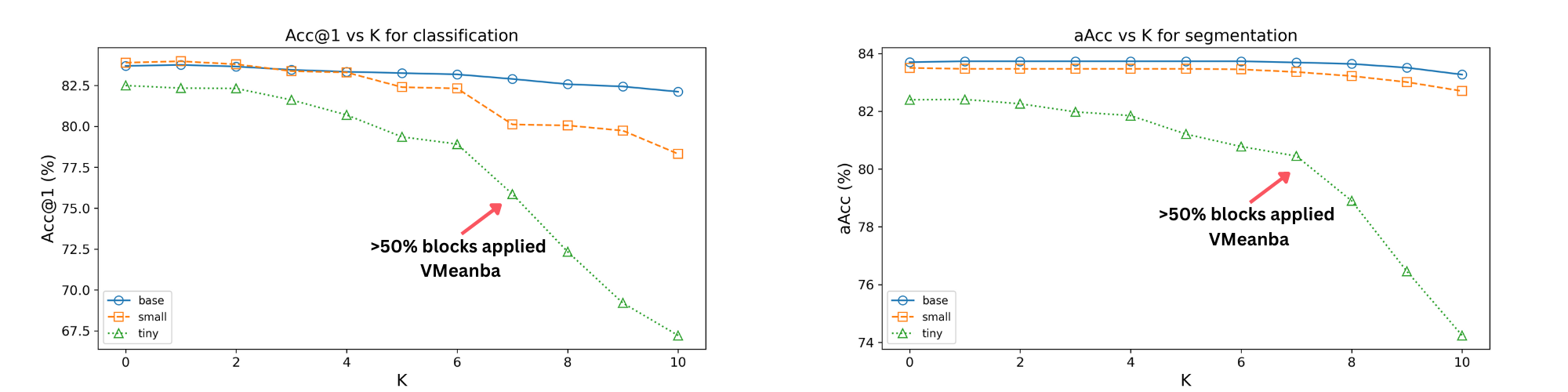}
    \caption{Accuracy versus K Analysis on classification and segmentation tasks by using \textit{VMeanba}. 
    This figure illustrates the trade-off between the value of $K$ and the associated accuracy drop. 
    By carefully selecting an appropriate $K$, the model's accuracy can be largely preserved.}
    \label{fig:4}
\end{figure*}

\subsection{Results on image classification and semantic segmentation}\label{subsection:results}
\textbf{Accuracy versus $K$ Analysis}. We applied the \textit{VMeanba} method to VMamba backbone models for both image classification and semantic segmentation tasks, varying the parameter $K$, as shown in Figure \ref{fig:4}. Our results indicate that the model accuracy remains largely unaffected when an appropriate $K$ value is chosen. However, there is a trade-off exists: increasing $K$ reduces inference time but leads to a more pronounced accuracy decline, as indicated by the arrows in the figure. Striking an optimal balance between accuracy and $K$ is essential. For example, selecting $K=10$ for the \texttt{base} model in image classification and semantic segmentation appears reasonable. In cases where accuracy drop is deemed unacceptable, one could still opt for a larger $K$ and retrain the model to recover performance. Since this study focuses on a training-free approach, retraining strategies are left for future work.

\begin{wraptable}{r}{5cm}
    \centering
    \begin{tabular}{ccc}
        \toprule
        Pruning Target & $K$ & Acc@1 \\
        \midrule
        \multirow{2}{*}{Linear Layers} & 0  & 83.5\% \\
        & 8 & 81.6\% \\
        \midrule
        \multirow{2}{*}{Conv2D Layers} & 0  & 80.1\% \\
        & 8 & 77.5\% \\
        \bottomrule
    \end{tabular}
    \caption{Accuracy comparison of \textit{VMeanba} with pruning on Linear and Conv2D layers using the \texttt{base} backbone.}
    \label{table:4}
\end{wraptable}

\subsection{Combined with Other Optimization Techniques} \label{subsection:combined}

We demonstrated that our \textit{VMeanba} method can be seamlessly integrated with other optimization techniques to enhance model efficiency. Specifically, we explored the effectiveness of combining \textit{VMeanba} with unstructured pruning on the VMamba \texttt{base} model 
for the image classification task using value $K = 8$. The results are summarized in Table \ref{table:4}. 
Pruning was applied to weight of linear layer or convolution 2D layer using the $l1$ norm, with a consistent pruning ratio of 40\%. 
Our findings indicate that the \textit{VMeanba} method is orthogonal to pruning, 
as it enhances efficiency while maintaining comparable accuracy, 
demonstrating that the two techniques can be combined without interference.
\section{Conclusion}\label{section:conclusion}
In this work, we introduced \texttt{VMeanba}, a novel, training-free model compression technique that reduces the inference time of the Mamba block in VMamba by applying a mean operation to reduce the dimensionality of input channel tensors in the associate scan operation. Our experimental results demonstrate that \texttt{VMeanba} enhances inference speed and throughput while maintaining competitive accuracy in VMamba.

This work contributes to the field by introducing a practical method for improving VMamba's efficiency and suggests future exploration of the dimensionality of input channel tensors and the kernel fusion of the discretization and selective scan operations to improve GPU utilization. Additionally, we envision extending \textit{VMeanba} to other computer vision tasks to evaluate its broader applicability and scalability.
\newpage

\appendix
\section{Preliminaries}\label{appendix:preliminaries}
In this section, we introduce some preliminaries of the State Space Model, SSM \cite{kalman1960new}, 
and two recently proposed methods using SSM, mainly selective state space model (Mamba)\cite{gu_mamba_2024} and VMamba\cite{liu_vmamba_2024}

\paragraph{State Space Model (SSM).} The SSM is a mathematical model that represents the evolution of a system over time. 
The model is specified as a set of equations that relate the state of the system to the observations at each time step. 
The most general form of the SSM is called continuous-time linear dynamical system, which is defined as equation (\ref{eq:1}).
\begin{equation} \label{eq:1}
    \begin{gathered}
        h'(t) = A(t)h(t) + B(t)u(t) \\
        y(t) = C(t)h(t) + D(t)u(t)
    \end{gathered}
\end{equation}
 $h(t) \in \mathbb{R}^n$ is the state variable at time step $t \in \mathbb{R}$, or usually called hidden variable in recent machine learning literature,
$u(t) \in \mathbb{R}^m$ is the input, $y(t) \in \mathbb{R}^p$ is the output, and $A(t) \in \mathbb{R}^{n \times n}$, $B(t) \in \mathbb{R}^{n \times m}$, 
$C(t) \in \mathbb{R}^{p \times m}$, $D(t) \in \mathbb{R}^{p \times m}$ are the system matrices at each time step. Note that in the following context, 
we treat $u(t)$ and $y(t)$ as scalars, i.e., $m = p = 1$.
The above continuous-time linear dynamical system can lead to a linear time-invariant (LTI) system when the system matrices $A(t)$, $B(t)$, $C(t)$, $D(t)$ are all time-invariant.
This LTI SSM then can be written as equation (\ref{eq:2}). It can be discretized into a discrete-time linear dynamical system, which is defined as equation (\ref{eq:3}). One of the frequent ways for this transformation utilized in the literature related to SSM is zero-order hold (ZOH) discretization, which is defined as equation (\ref{eq:4}). Besides, it can further written as a convolution form (\ref{eq:5}).
\begin{equation} \label{eq:2}
    \begin{gathered}
        h'(t) = Ah(t) + Bu(t) \\
        y(t) = Ch(t) + Du(t)
    \end{gathered}
\end{equation}

\begin{equation} \label{eq:3}
    \begin{gathered}
        h_t = \bar{A}h_{t-1} + \bar{B}u_t \\
        y_t = Ch_t + Du_t
    \end{gathered}
\end{equation}
\begin{equation}\label{eq:4}
    \begin{gathered}
        \bar{A} = \exp(\Delta A) \\
        \bar{B} = (\Delta A)^{-1}\exp(\Delta A -  I)\Delta B
    \end{gathered}
\end{equation}
\begin{equation} \label{eq:5}
    \begin{gathered}
        \bar{K} = (C\bar{B}, C\bar{A}\bar{B}, ..., C\bar{A}^k\bar{B},  ...) \\
        y = x \ast \bar{K}
    \end{gathered}
\end{equation}
\textbf{Selective State Space Model (Mamba)}. Mamba is the discrete-time linear dynamical system with a timescale parameter $\Delta$ that transforms the continuous variables $A$, $B$ to discrete variables $\bar{A}$, $\bar{B}$.
In addition to discretization, Mamba also relax the time-invariant constraint of the system matrices by introducing \textsl{selection} mechanism, which simply makes several parameters
$\Delta$, $B$, $C$ to be time-varying by functions $s$ of the input $u$. Specifically defined as equation (\ref{eq:6}).
\begin{equation}\label{eq:6}
    \begin{gathered}
        s_B(u) = Linear_N(u) \\
        s_C(u) = Linear_N(u) \\
        s_{\Delta}(u) = Broadcast_{D}(Linear_{1}(u)) \\
        \Delta = \tau_{\Delta}(Parameter + s_{\Delta}(u))
    \end{gathered}
\end{equation}
The $Linear_d$ is a parameterized linear projection to dimension $d$, and the $\tau_{\Delta} = softplus$. As the selection mechanism loses the equivalence to convolution form (\ref{eq:4}),
to avoid the sequential recurrence, Mamba further incorporates a work-efficient parallel algorithm, \textit{associate scan}, into its GPU kernel implementation to facilitate parallel computation of the system.
\paragraph{VMamba} The original Mamba block is designed for 1-dimensional input and output, which is not suitable for computer vision tasks. VMamba proposed a new module called 2D-Selective-Scan (SS2D)
for adapting Mamba to 2D input and output. The SS2D module is composed of three steps: cross-scan, selective scan (Mamba block), and cross merge. The cross-scan unfold the input feature map along
four directions, forming 4 sets of 1D sequences. Then the selective scan processes each 1D sequence in parallel. The cross-merge finally merges the 4 sets of 1D sequences back to 2D feature map. The
cross-scan and cross-merge are called Cross Scan Module (CSM) together, and by this way, the model can have a global receptive field. VMamba further stack multiple SS2D blocks in a layer, and then stack
 layers to form the whole model.
\section{Computation Complexity}\label{appendix:computation-complexity}
\paragraph{Complexity of SSM} The computational complexity of the associated scan operation in Mamba block, measured in floating-point operations (FLOPs), 
is derived from processing a sequence of length $L$, which requires $2L$ operations. Furthermore, 
the input to the scan operation incurs an additional cost of 3 FLOPs, leading to a total of $ 3 \times 2BLD$, 
where $B$ is the batch size, $L$ is the sequence length, and $D$ is the inner dimension. 

In the context of the 
SSM system, computations involve multiplications for $\bar{B}u_t$ and $Ch_t$, which amount to $2BLD$, 
and additions for $Ch_t$ and $D$, totaling $BLD$ FLOPs. Consequently, the overall FLOPs for the SSM system is $3BLD$. The total FLOPs for the Mamba block, therefore, aggregate to $3 \times 2BLD + 3BLD$.

\paragraph{Complexity of reduced SSM} The reduction in FLOPs can be achieved by employing the $I_{basis}$, which consists of $9BLd$ FLOPs,
and additional FLOPs for the reduce operation and broadcast operation. The total reduction in FLOPs is 
summarized by the equation (\ref{eq:8}):

\begin{equation}\label{eq:8}
    \begin{aligned}
        FLOP_{Mamba} &= FLOP_{scan} + FLOP_{SSM} \\
        FLOP_{original} &= 3 \times 2BLD + 3BLD \\
        FLOP_{reduction} &= 9BLd + FLOP_{reduce\_op} \\ 
        \;&+ FLOP_{broadcast}
    \end{aligned}
\end{equation}

\paragraph{Complexity of \textit{VMeanba}} The mean operator contribute only $BLD + BL$ FLOPs, and the broadcast operator is just a memory operation. The reduced FLOPs is then $B(10 + D)L$ FLOPs, comparing to the original $9BDL$ FLOPs,
we achieve 89\% FLOPs reduction ($10 << D$).

\section{Algorithm}\label{appendix:algorithm}
\begin{algorithm} 
    \caption{VMeanba Layer Selection Pipeline}
    \textbf{Input}: $Model$, $D_{val}$, $K$, $CalculateScore$ \\
    \textbf{Output}: $layersToApply$ 
    \begin{algorithmic}[1]
        \STATE $Scores \leftarrow [\;]$
        \FOR{$layer$ in $Layers$}
            \STATE $s \leftarrow CalculateScore(layer, Model, D_{val})$
            \STATE $Scores \leftarrow Scores + s$
        \ENDFOR
        \STATE $Layers \leftarrow \text{Sort}(S_{layer})$
        \STATE $layersToApply \leftarrow Layers[:K]$
        \STATE \textbf{return} $layersToApply$
    \end{algorithmic}
\end{algorithm}

\section{Experiments Setup}\label{appendix:exp-setup}
\textbf{Datasets}. The datasets we use for our \textit{VMeanba} experiments are the ImageNet-1k dataset \cite{deng2009imagenet} for image classification and 
the ADE20k dataset \cite{zhou2017scene} for semantic segmentation. We only use the validation set of them for the experiments. 
The ImageNet-1k dataset contains 50k validation images from 1k classes, and the ADE20k dataset contains 2k images for validation, with pixel-level annotations.\\
\textbf{Models}. We use the VMamba pre-trained backbone models \cite{liu_vmamba_2024} for both tasks. The backbone models is first trained on the ImageNet-1k training dataset. 
It is then used as the pre-trained backbone models for downstream task. The segmentation task use the UperNet \cite{xiao2018unified} on top of the VMamba pre-trained backbone models, 
and trained on the ADE20k training dataset. The VMamba backbone models have three different versions: {\it tiny}, {\it small}, and {\it base}. There are two mainly differences between 
these versions: the number of layers and the dimension of the $L$ and $D$ in the SS2D block. All of the backbone models have four layers and the {\it tiny} 
version is stack as [2, 2, 8, 2], while the other two versions are stack as [2, 2, 20, 2]. The dimension of the $L$ and $D$ is different across 
two tasks, both of them remain the same inside each layer. However, the dimension of the $D$ grows by a factor of 2,  and the dimension of the $L$ 
scale down by a factor of 4 along the layers.\\
\textbf{Kernel Implementation}. The original CUDA kernel for the Mamba block includes both the discretization and scan operations, 
dividing the GPU multiprocessor into a 2D grid blocks based on the batch size and inner dimension. 
In this configuration, multiple threads within the block handle the scan operation. 
However, since the discretization process is not the focus of this study, and the original approach of dividing the inner dimension 
across blocks is not compatible with our \textit{VMeanba} method, we developed a new CUDA kernel. 
This new kernel exclusively handles the scan operation, with the discretization process executed outside the kernel. 
All experiments conducted in this paper are based on this optimized kernel. 
Future work includes integrating the discretization and scan operations into a single kernel for further optimization.\\
\textbf{Additional Information}. The evaluation metric for the image classification task is top-1 accuracy, 
while for the semantic segmentation task, we utilized all pixel accuracy (aAcc). 
The batch size for the image classification task is set to 128, whereas for the semantic segmentation task, 
it is limited to 1 due to the dynamic input size present in the validation set. 
All experiments were conducted on a single NVIDIA RTX A6000 GPU with 48GB of memory. 
The profiling was performed using NVTX API, Nvidia Nsight Systems, and Nvidia Nsight Compute tools.

\section{More Experiment results}\label{appendix:more-exp}
\begin{table*}[ht!]
    \centering
    \caption{Speedup analysis of the \textit{VMeanba} method compared to the original inner dimension size kernel. All VMeanba times are approximately 0.02 ms.\\}
    \begin{tabular}{cccccc}
        \toprule
        Backbone & Inner dimension & Sequence length & Original time (ms) & Speedup \\
        \midrule
        \multirow{4}{*}{Tiny \& Small} 
        & 384  & 3136 & 5.46 & 273x \\
        & 768  & 784  & 2.23 & 112x \\
        & 1536 & 196  & 1.10 & 55x \\
        & 3072 & 49   & 0.71 & 36x \\
        \midrule
        \multirow{4}{*}{Base} 
        & 512  & 3136 & 5.86 & 293x \\
        & 1024 & 784  & 2.94 & 147x \\
        & 2048 & 196  & 1.47 & 74x \\
        & 4096 & 49   & 0.93 & 47x \\
        \bottomrule
    \end{tabular}
    \label{table:1}
\end{table*}
\begin{table*}[ht!]
    \caption{GPU kernel memory usage with and without the \textit{VMeanba} method. \dag~indicates that the original kernel memory usage is too small to be measured.\\}
    \centering
\begin{tabular}{>{\centering\arraybackslash}p{3cm} >{\centering\arraybackslash}p{3cm} >{\centering\arraybackslash}p{3cm} >{\centering\arraybackslash}p{3cm}}
        \toprule
    Inner dimension & Sequence length & Original memory R/W (Bytes) & Optimized memory R/W (Bytes) \\
        \midrule
        512  & 3136 & 3.3G / 823.5M & 6.4M / 1.3M \\
        1024 & 784  & 1.6G / 411.9M & 1.6M / 14.5K \\
        2048 & 196  & 822.1M / 207.5M & 412.4K / 5.8K \\
        4096 & 49   & 411.1M / 107.5M & 108.5K / 0\dag \\
        \bottomrule
    \end{tabular}
    \label{table:2}
\end{table*}
\textbf{Kernel Analysis}
We analyzed GPU kernel speedup and memory usage when applying \textit{VMeanba} across varying scan sequence lengths and inner dimensions, as shown in Tables \ref{table:1} and \ref{table:2}. Optimized kernel times, consistently around 0.02 ms, are excluded from Table \ref{table:1}. The \textit{VMeanba} method achieves up to 293x speedup, particularly for longer scan sequences, aligning with the $O(DL)$ complexity discussed in \ref{appendix:computation-complexity}. Additionally, memory transfer between global and shared memory is significantly reduced, enabling longer scan sequences and larger batch sizes for improved throughput.

\newpage
\begin{table}[ht]
     \caption{Batch inference time comparison for the VMamba models with and without the \textit{VMeanba} method on the image classification task.}
    \centering
    \begin{tabular}{ccccc}
        \toprule
        Backbone & $K$ & Accuracy (Acc@1 / aAcc) & Batch Inference Time (ms) & Speedup \\
        \midrule
        \multirow{4}{*}{Tiny} & 0 & 82.5\% & 283 & 1x \\
         & 2 & 82.3\% & 261 & 1.08x \\
         & 4 & 80.7\% & 252 & 1.12x \\
         & 8 & 72.3\% & 240 & 1.18x \\
        \midrule
        \multirow{4}{*}{Small} & 0 & 83.9\% & 415 & 1x \\
         & 2 & 83.8\% & 393 & 1.06x \\
         & 4 & 83.3\% & 391 & 1.06x \\
         & 8 & 80.1\% & 383 & 1.08x \\
        \midrule
        \multirow{4}{*}{Base} & 0 & 83.7\% & 527 & 1x \\
         & 2 & 83.7\% & 519 & 1.02x \\
         & 4 & 83.3\% & 515 & 1.02x \\
         & 8 & 82.6\% & 508 & 1.04x \\
        \bottomrule
    \end{tabular}
    \label{table:3}
\end{table}
\textbf{Batch Inference Time Analysis}. We compared batch inference times of VMamba models with and without the proposed \textit{VMeanba} method across three backbone models on an image classification task (Table \ref{table:3}). The application of \textit{VMeanba} reduced inference times, increasingly so as the value of $K$ increased due to time savings from applying the mean operation to more layers. Notably, the \texttt{base} model exhibited less speedup compared to the \texttt{small} and \texttt{tiny} models, likely due to its larger inner dimension size incurring greater time consumption during discretization and the mean operation.

\bibliographystyle{abbrv}
\bibliography{neurips_2024}
\end{document}